\documentclass{ifacconf}
\usepackage{algorithm}
\usepackage{algpseudocode}
\usepackage{graphicx}      
\usepackage{booktabs}
\usepackage{tabularx}
\usepackage{enumitem}
\usepackage{amsmath}
\usepackage{multirow}
\usepackage{amssymb}
\usepackage{subcaption}

\usepackage{natbib}        
\usepackage{indentfirst} 
\begin{document}
\begin{frontmatter}

\title{Evaluating Path Planning Strategies for Efficient Nitrate Sampling in Crop Rows} 


\author[First]{Ruiji Liu} 
\author[First]{Abigail Breitfeld} 
\author[First]{Srinivasan Vijayarangan} 
\author[First]{George Kantor}
\author[First]{Francisco Yandun}

\address[First]{Carnegie Mellon University, 
   Pittsburgh, PA 15213 USA (e-mail: ruijil@andrew.cmu.edu , abreitfe@andrew.cmu.edu , svijaya1@andrew.cmu.edu , gkantor@andrew.cmu.edu , fyandun@andrew.cmu.edu).}

\begin{abstract}                
This paper presents a pipeline that combines high-resolution orthomosaic maps generated from UAS imagery with GPS-based global navigation to guide a skid-steered ground robot. We evaluated three path planning strategies: A* Graph search, Deep Q-learning (DQN) model, and Heuristic search, benchmarking them on planning time and success rate in realistic simulation environments. Experimental results reveal that the Heuristic search achieves the fastest planning times (0.28\,ms) and a 100\% success rate, while the A* approach delivers near-optimal performance, and the DQN model, despite its adaptability, incurs longer planning delays and occasional suboptimal routing. These results highlight the advantages of deterministic rule-based methods in geometrically constrained crop-row environments and lay the groundwork for future hybrid strategies in precision agriculture.
\end{abstract}


\end{frontmatter}

\section{Introduction}
\label{sec:intro}
\vspace*{-4pt}
Autonomous navigation in agricultural fields is challenging due to structured layouts with unstructured variability. Key challenges include (1) visual clutter from vegetation; (2) narrow inter-row spacing, allowing little margin for error; and (3) minimal global references, as repeating field patterns hinder reliable localization. 
We used the skid-steered Amiga robot, developed by Farm-ng, with a side-mounted robotic arm for manipulation tasks. It straddles crop rows for pre-planned missions with predefined waypoints. However, dynamic tasks like nitrate sampling in cornstalks require real-time path planning based on prior measurements \citep{corn}.

\begin{figure}
\begin{center}
\includegraphics[width=0.6\columnwidth]{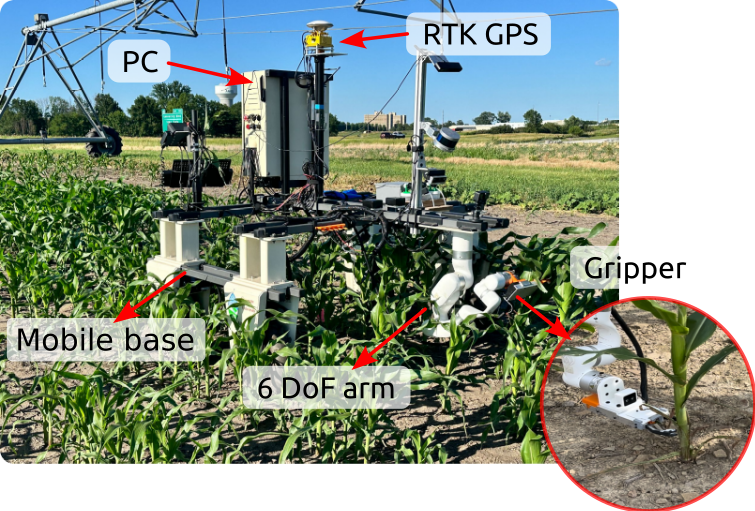}    
\caption{Robot designed for sampling nitrate content in corn fields. It features a robotic arm equipped with a custom gripper for nitrate probing, navigation sensors, and an onboard computer.} 
\label{fig:amiga}
\end{center}
\end{figure}

In this paper, we present a benchmarking study of path planning strategies for efficiently sampling nitrate content in cornstalks using the Amiga robot, shown in Fig. \ref{fig:amiga}. Although our test case focuses on the sampling of nitrates in cornstalks, the planners discussed in this work can be generalized to other agricultural tasks that involve navigating crop rows and collecting samples from plants.

It is important to note that our study does not address the optimal ordering of sampling points for full coverage (similar to the Traveling Salesman Problem). Instead, we focus on evaluating planning strategies that generate the shortest possible path while navigating between rows and taking samples, satisfying general and specific constraints. In our application, additionally to the regular planning constraints (e.g., presence of obstacles), this problem requires the robot to reach the goal with a certain orientation with respect to the rows (see Section \ref{sec:controller}). We evaluated three methods and report its performance in terms of planning time and success rate.

\section{Related Work}
Global food demand is expected to rise by 29\% to 91\% by 2050, requiring increased agricultural productivity \citep{mouel17}. Autonomous navigation in agricultural fields is crucial for improving efficiency and reducing labor dependency. UAV-based mapping has been proposed for row crop navigation \citep{mansur25}, but its effectiveness declines as crop canopies grow, obscuring ground visibility, especially in crops like soybeans.

Vision-based navigation methods have been explored, including camera-based field navigation for seed sowing \citep{santhi17}, texture tracking for crop row detection without relying on color or spacing \citep{english14}, and segmentation-based crop row detection in maize fields (Guerrero et al., 2013). While effective in controlled conditions, vision-based systems are highly sensitive to lighting changes and occlusions. Even small deviations from the crop row can cause navigation failures due to the lack of global context.

LiDAR-based approaches offer an alternative. Autonomous navigation in strawberry fields has been achieved by integrating waypoint navigation with LiDAR-based bed detection \citep{fujinaga25}, and LiDAR-based navigation has been demonstrated in cornfields \citep{kim25}. While resilient to lighting variations, LiDAR struggles with reconstructing dense plant structures due to point density limitations. These challenges underscore the need for more robust navigation strategies that can operate reliably in real-world agricultural environments. Our work is similar to the GPS based methods of \cite{moeller20} and \cite{shaik18}. 

For dynamic planning within crop rows, various approaches can be employed, ranging from simple heuristics that navigate to the nearest goal-aligned row to classic search methods like A* \citep{santos2020path} for improved decision-making. Given that the robot’s maneuverability depends on the arm’s mounting side, hybrid A* \citep{Kurzer1057261} offers a more suitable approach. However, traditional search algorithms struggle with large maps.  Custom planners that encode exploration actions differently or leverage data-driven approaches, such as reinforcement learning \citep{singh2023review}, offer viable alternatives for efficient planning. This paper analyzes the differences between these methods and presents the results.
\section{Global mapping and navigation}
\begin{figure}
\begin{center}
\includegraphics[width=0.8\columnwidth]{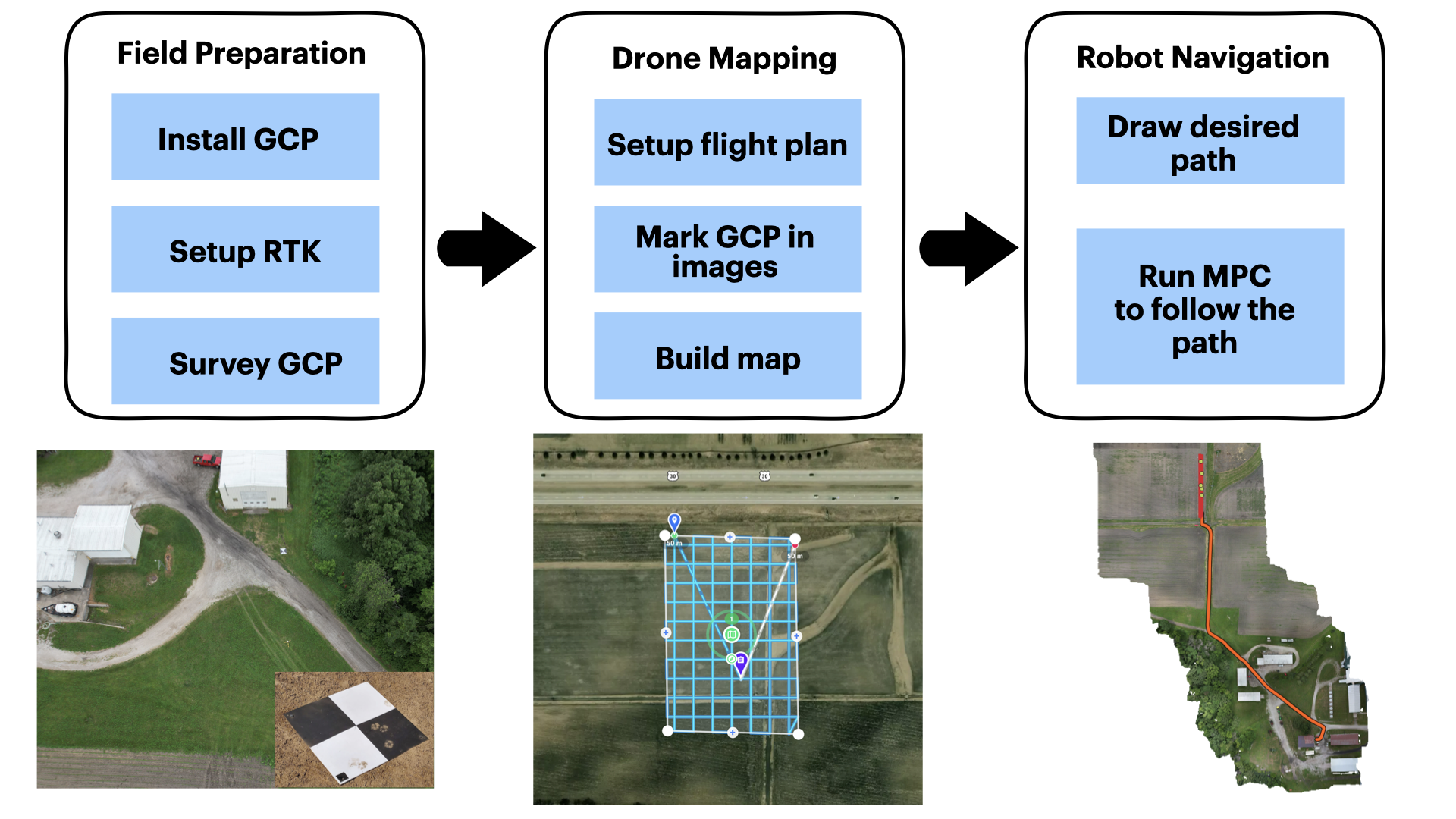}    
\caption{Global navigation layout. A pre-computed map of the field is used to define the ground robot's path. Then it tracks it to reach designated nitrate sampling points.} 
\label{fig:global_nav_arch}
\end{center}
\end{figure}

Our navigation strategy follows a two-step approach. First, an unmanned aerial system (UAS) surveys the area where the ground robot will operate. Using the collected aerial imagery, we generate an orthomosaic map offline, where a user can define the path the ground robot will follow. We named this approach ``global" navigation because the ground robot only relies on the pre-built map and the user input to autonomously navigate in the field. For the experiments presented in this paper, the robot does not utilize local perception when driving between rows to adjust its actions. Fig. \ref{fig:global_nav_arch} depicts the overall approach and the following Sections describe each of the steps in the global navigation strategy.

\label{sec:global_nv}

\subsection{Aerial Survey}
For the ground robot to navigate accurately in any crop field, the aerial map needs to be geo-referenced with RTK-level accuracy (i.e., centimeter-level). Our method does not use an RTK GPS mounted on the UAS; instead, the images captured by the drone during the flight are geo-tagged using its built-in GPS. To generate a highly accurate map, we placed several ground control points (GCPs) in the field and survey their positions with a ground-based RTK system. Based on our experience, the number of GCPs required depends on the field size and terrain. Larger fields (greater than 4 hectares) typically require 3–4 GCPs, but if the terrain includes slopes or significant elevation changes, additional GCPs may be necessary. To combine the drone imagery and the surveyed GCP locations, we used a commercial photogrammetry software (Agisoft Metashape), generating a highly accurate map of the field. Through extensive field tests, we found that a map resolution of at least 10cm/pixel provides the best accuracy for the ground robot to follow narrow crop rows effectively.

The final map is geo-referenced in UTM coordinates and exported as a GeoTIFF file, which can be opened in any Geographic Information System (GIS) software. Since our test cases involves the robot navigating parallel to the crop rows, we drew straight lines in the traversable regions of the rows and use built-in GIS features to automatically generate parallel lines in the traversable regions of the rows. These lines are discretized into waypoints, and together with sampling locations (which can be generated automatically or selected manually), are loaded onto the robot’s onboard computer for execution.
\subsection{Ground Robot Controller}
\label{sec:controller}
The robot's control system relies on an accurate localization setup to track the planned path. It used an RTK-GPS receiver mounted on the robot, wheel encoders, and an onboard IMU (Inertial Measurement Unit), whose measurements are fused using an Extended Kalman Filter. The main task of the control system was to drive the robot through the rows, stop at the sampling points, and navigate between rows if necessary. 

We implemented a Model Predictive Controller (MPC) to track the generated waypoints while navigating between rows, and we adopted the approach described in \citep{ruiji2025} for row-switching. Specifically, the robot utilized the MPC to follow the rows and transitioned to a PID controller for exiting a row, executing a turn, driving straight, and making a final turn to enter the next row. Additionally, a preprocessing stage was used to smooth the path and calculate curvatures and speed profile for the global route. 

After reading the global plan and the sampling points, a local path needs to be generated from the current robot position to the next sampling location. While the optimal order of the points could be formulated as a Traveling Salesman Problem, this was not addressed in this work. Instead, our focus is on the \textit{local planner}, since for each sampling locations, the robot needs to find an optimal path that saves time and energy.

\section{Planning Problem}
\label{sec:planning_problem}
As mentioned in Section \ref{sec:intro}, the overall goal of the robotic system is to autonomously sample the nitrate content of corn plants in large fields. More formally: Let $\mathcal{C}$ be the configuration space of the robot. Given boundary conditions $q_{start} \subseteq \mathcal{C}_{free}, q_{goal} \subseteq \mathcal{C}_{free}$ and a map $\mathcal{M}:\mathbb{Z}^2\rightarrow \{0,1\}$, where
\begin{align}
    \mathcal{C}_{free} = \{(x,y,\theta) | M(x,y) = 0 \} \\
    \mathcal{C}_{obst} = \{(x,y,\theta) | M(x,y) = 1 \} 
\end{align}
Are the free space where the robot can navigate and the space that indicates the presence of plants, respectively. we aim to find a feasible and optimal path $\rho:[0,1]\rightarrow \mathcal{C}_{free}$ for moving from the initial configuration to the sampling point (i.e., $q_{goal}$) while avoiding the crops, minimizing the total distance traveled by the robot. To clarify the terminology used in the following sections, we define rows as the planted areas and corridors as the drivable spaces between them.

This paper focuses on the evaluating the strategies that a \textit{local planner} can use to find a path $\rho(0,1)$ that efficiently and effectively reach sampling points. An additional constraint in the planning problem originates from the robotic arm: it is mounted in a way that can only sample plants located on its left side. This restriction influences the feasible paths the robot can take when approaching a sampling point. However, the robot controller supports both forward and backward motion, which allows for more efficient maneuvering within the crop field.

Options for solving this task in the state of the art range from classical methods like Dijkstra or A* \citep{santos2020path} to newer reinforcement learning approaches \citep{singh2023review}. In this work we implemented and compared four known methods from the state of the art, namely: A* search, DQN-based planner, and a heuristic-based approach. In this way, we can compare and evaluate the suitability and the advantages or disadvantages of each one. The specific formulation of each method for our particular problem is discussed in the following Sections.

\subsection{Graph Search Formulation}
\label{sec:graph_search}

\begin{figure}[t]
    \centering
    \includegraphics[width =0.35\textwidth]{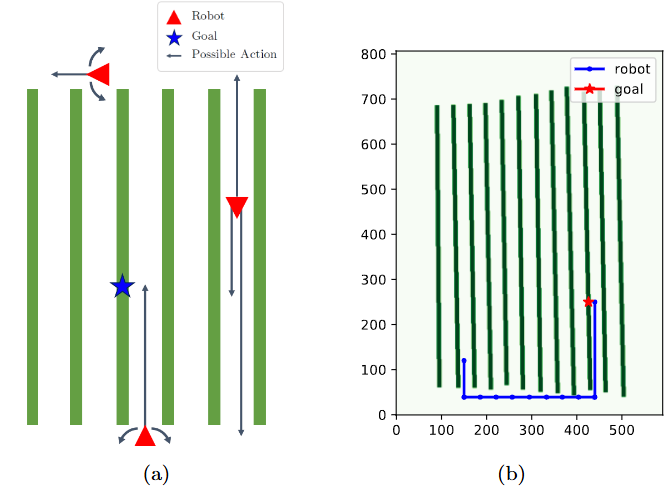}
    \caption{Overview of A* Graph search: (a) possible actions based on the robot’s position in the field, and (b) an example planned trajectory.}
    \label{fig:A_star}
\end{figure}
Our Graph search formulation uses the A* approach \citep{hart1968formal} to plan a path between an initial position and orientation $[x_1, y_1, \phi]$ to a target sampling point $[x_2, y_2]$. Note the final orientation is not specified as the robot can reach a sampling point from two sides of the rows, with different orientations. Due to the large size of some fields, we do not explicitly represent the graph using an occupancy grid that allows the algorithm to explore the complete map. Instead, as shown in Fig. \ref{fig:A_star}, we implicitly build a graph representation based on the robot's position, orientation, and the natural constraints of the environment (i.e., the robot drives parallel to crop rows and moves between them only at the ends). We assume obstacles (rows) are oriented top to bottom, and if the map is not aligned this way, a preprocessing step rotates it for consistency.

Based on the previous configuration, shown in Fig. \ref{fig:A_star}, we first determined the x-coordinates of each corridor. The row containing the initial position $x_1$ is found by identifying the closest x-coordinate. Once the starting row is known, the algorithm begins. At any given position, the robot can take one of three actions, depending on whether it is within a corridor or at the ends of the rows. If the robot is within a corridor, the possible actions are: i) Move forward/backward to the top of the current corridor, ii) Move forward/backward to the bottom of the current corridor and iii) Move forward/backward to the target's y-coordinate ($y_2$) if it is in the current corridor.
Conversely, if the robot is at the top or bottom edge of a row, the possible actions are: i) turn left, ii) turn right, and iii) move forward/backward, depending on whether the robot is at the top or bottom edge. 

Given that the robot track spans various rows, we forbade the actions of turning left or right within a corridor to avoid trampling crops underneath the chassis. As opposed to classic A*, which performs the search using an occupancy grid map, our formulation uses the structure of the environment to reduce the number of node expansions performed. Instead of planning a path one grid location at a time, our search must only consider the actions enumerated above, which leads to a significant time speedup.

The planner considers the target location to be effectively two different goal positions. The robot can reach the target from the right side, orienting towards the top of the map, or from the left, facing the bottom of the map.
The planner takes these two possible orientations into account when determining if the robot has reached the goal. Given this set of actions and goal states, the search proceeds following the A* procedure.


\begin{figure}[t]
    \centering
    \includegraphics[width =0.3\textwidth]{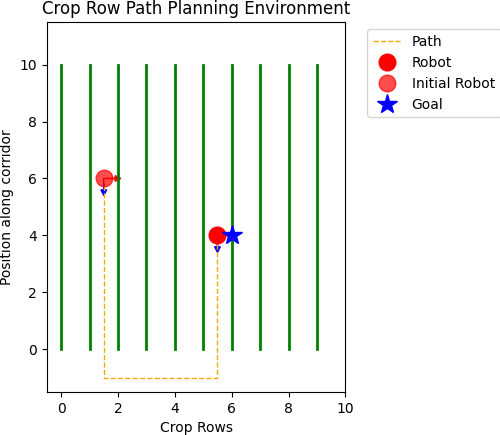}
    \caption{Custom Gym environment (\textit{CropRowEnv}) for crop row path planning RL agent training (10 crop rows x 10 units length), with a two-component action space. }
    \label{fig:env}
\end{figure}
\subsection{Reinforcement Learning Formulation} 
\label{sec:RL}
We trained a reinforcement learning (RL) agent to find the shortest path to the sampling point. The simulation environment employed during training was implemented within the OpenAI gym framework \citep{gymopenai}. As shown in Fig.\ref{fig:env}, it featured rows spaced evenly at 1 unit with a length of 10 units, and the number of rows could be adjusted to match the field size where the robot will operate. As explained in Section \ref{sec:dqn}, this set-up, with a constant corridor length, enabled the deployment of the policy in fields with a specific number of rows ($\leq$ rows used in training) and arbitrary row lengths. However, the bigger the number of crop rows, the longer the training time. 

As discussed in Section~\ref{sec:planning_problem}, a key constraint is that the robot arm's work zone is exclusively on its left side. To simulate this constraint, the blue arrow indicates the robot's orientation, while the red arrow denotes the robot's work zone, as shown in Fig.\ref{fig:env}. A goal is considered reached only when the sampling point lies within this work zone. During training, we encoded the robot movement by restricting the RL agent action space as a two value vector:
\begin{itemize}
    \item The first element specifies the orientation of the robot relative to the map (0 for upward, 1 for downward).
    \item The second element controls the movement command:
    \begin{itemize}
        \item If $<$ 2, it is a vertical move (0 = forward, 1 = backward).
        \item If $\geq$ 2, it indicates a corridor-switching action (target corridor = value - 1.5).
    \end{itemize}
\end{itemize}
The observation space consisted of the position of the robot, its orientation, and the position of the goal $[x_r, y_r, \theta_r, g_x, g_y]$.

An important aspect of the RL framework is the design of the reward function, which plays a key role in guiding the agent toward the optimal solution. In our environment, the agent incurs a small penalty of $-0.2$ for each step taken within the corridor, and a penalty of $-0.2 \times \text{number of traversed\_rows}$ when switching corridors, encouraging the agent to find the shortest path. A larger penalty of $-1.5$ is applied for undesirable behaviors, such as turning or oscillating back and forth within the corridors. This helps discourage inefficient movements and encourages more purposeful navigation. Upon reaching the goal, the agent is rewarded with a significant reward of $+20$, reinforcing the correct behavior and the completion of the task. Additionally, to encourage efficient path selection, the agent is awarded a small bonus of $+5$ if it chooses the closer corridor based on its initial position, since the goal can be reached from either adjacent corridor. 



\subsubsection{DQN Model}\label{sec:dqn}
We employed the Deep Q-Learning (DQN) implementation from Stable Baselines3 \citep{stable-baselines3}, using a three-layer MLP (1024 neurons) with ReLU non-linearities to train our RL agent. The training process comprises two stages: rollout and update. The rollout stage runs for $100,000$ time steps, in which the agent interacts with the environment, initially selecting actions in a completely random manner to encourage exploration and collecting experiences (i.e., observations, rewards, actions, and terminal statuses) that are stored in a replay buffer. Subsequently, the update stage samples mini-batches from the buffer and trains the neural network by minimizing the difference between predicted and target Q-values. The target Q-value is computed using the Bellman equation \citep{bellman}. 

DQN stabilizes training by using a target network, a periodic copy of the Q-network. This target network calculates the target Q-value, preventing instability from continuously updating the same Q-network. By decoupling the target Q-value, this mechanism reduces large, erratic updates, ensuring more stable learning. These intrinsic learning mechanism enables the agent to refine its internal estimation of the optimal action-value function, gradually reducing its exploratory behavior and converging toward a policy that maximizes cumulative rewards.


Directly training the agent in an environment with a large number of crop rows produces highly sparse positive rewards, resulting in fewer successful episodes and slower convergence. To address this issue, we applied \textit{curriculum learning}: training agents in simpler environments and then fine tuning the weights of the neural network by gradually increasing the environment complexity. In our case, we started training the model in an environment with 5 crop rows and a corridor length of 10 units, then increased the number of crop rows by 5, eventually training in an environment with 65 crop rows and the same 10 unit corridor length, since this scenario corresponds to the largest field we encountered in real agricultural settings (with a corridor length of 200\,m). To reduce the training variability as the number of crop rows increases, the step count was consistently maintained at 10 per corridor. In addition, the duration of the training was gradually extended as the environment grew in complexity. Ultimately, the trained model outputs a sequence of actions in the form of $[Orientation, Move\_action]$, guiding the robot toward the goal. When deploying the policy in the real field, the length of the corridor was segmented into 10 sections, corresponding to the 10 units of corridor length during training. The initial robot and goal positions were assigned to the nearest sections.

\subsection{Heuristic Search}\label{sec:heuristic}
Within the same customized OpenAI Gym environment, we implemented a heuristic method to solve the same path planning problem. As detailed in Algorithm~\ref{alg:Heuristic}, the method similar to the RL agent, accepts as input the robot’s initial position and orientation $[x_1, y_1, \phi]$, the coordinates of the sampling point $[x_2, y_2]$, and the environmental dimensions (i.e., number of crop rows and corridor length). The method outputs a sequence of action vectors, $[Orientation, Move\_action]$.

\begin{algorithm}[t]
\caption{Heuristic Path Planning in Crop Rows}\label{alg:Heuristic}
\small 
\begin{algorithmic}[1]
\State $Env \gets (Ini\_pos, Goal, Env\_dimension)$
\State $Now\_pos \gets Env$
\State Compute minimal vertical path and exit direction
\If{Initial corridor X $<$ Goal corridor X}
    \State $Target\_Corridor \gets Goal\_Corridor - 0.5$
\Else
    \State $Target\_Corridor \gets Goal\_Corridor + 0.5$
\EndIf
\State Compute action sequence [Orientation, Move action]
\State \textbf{Post-process:} Remove duplicates and output the action sequence
\end{algorithmic}
\end{algorithm}

\begin{figure}[b]
    \centering
    \includegraphics[width =0.48\textwidth]{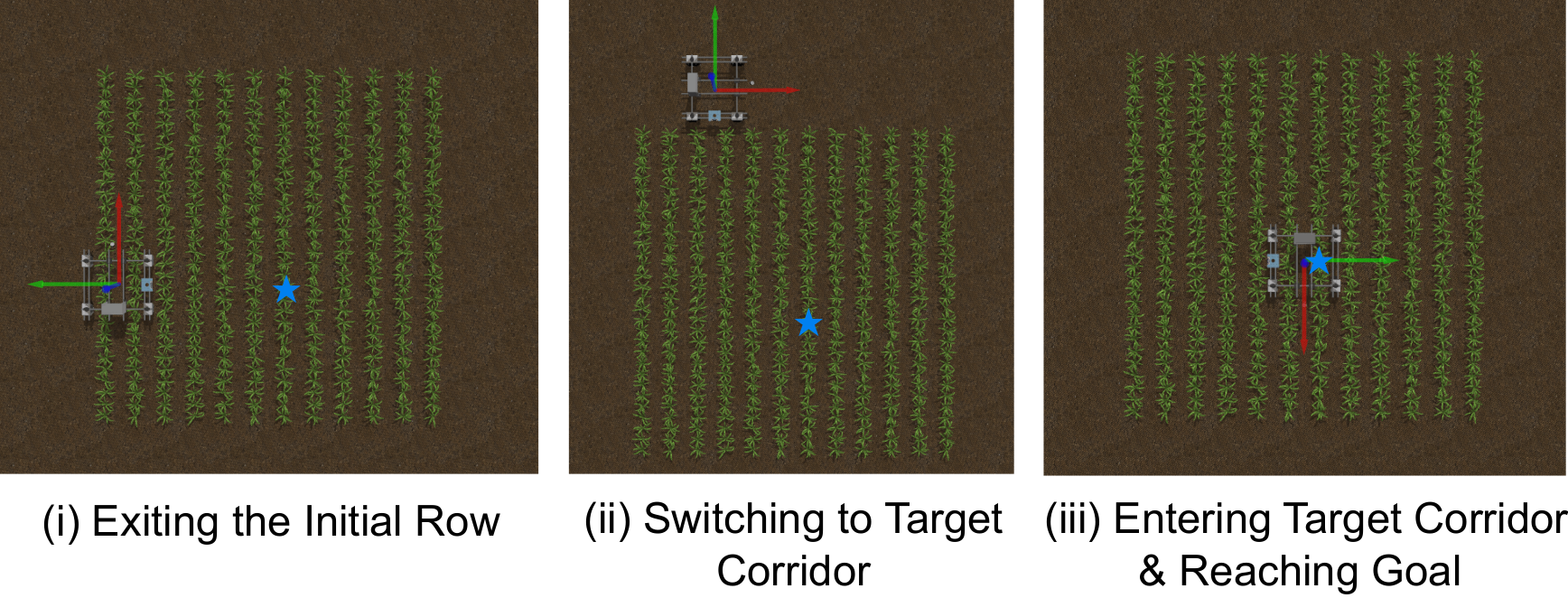}
    \caption{Example deployment of output actions [0,0], [1,7], [1,0] in a simulated field. The blue star represents the goal sampling point.}
    \label{fig:sim}
\end{figure}
The path planning problem is decomposed into three stages: (i) navigating and exiting the initial corridor, (ii) switching to the target corridor (if required), and (iii) entering and navigating within the target corridor (if required). Initially, based on the environmental dimensions, the starting position of the robot, and the sampling point, the method determines whether the robot should exit the upper or lower edge or proceed directly to reach the goal in the current corridor. If the robot reaches the end of the corridor without reaching the goal, the target corridor is identified according to the relative positions of the robot and the sampling point. Since the sampling point can be accessed from either of the adjacent corridors, the robot selects the one that minimizes the travel distance. Once the corridor switch is completed, the robot enters the target corridor according to its current edge position. Finally, the method determines the orientation and movement action in each step, removes redundant actions, and outputs a sequence of action vectors in the form of $[Orientation, Move\_action]$.

\section{Experimental Results}
Since the goal of this paper is to benchmark the planning strategies described before, we do not present evaluations on the performance of the global navigation strategy described in Section \ref{sec:global_nv}. However, the performance of the controller in following crop rows in other tasks has been reported in \cite{silwal2022bumblebee, ruiji2025}.

\begin{table}[t]
\small 
\caption{Performance Comparison of Various Path Planning Algorithms.}
\label{tab:tab1}
\centering
\begin{tabularx}{0.45\textwidth}{
  >{\centering\arraybackslash}X
  >{\centering\arraybackslash}X
  >{\centering\arraybackslash}m{1.8cm} 
}
\hline
Method & Avg. Planning Time & Success Rate \\
\hline
\hline
\addlinespace[0.3ex] 
DQN model & 2.78 ms & 96.33\% \\
Heuristic Search & 0.28 ms & 100\% \\
Graph Search & 1.40 ms & 99.13\% \\
\addlinespace[0.3ex]
\hline
\end{tabularx}
\end{table}

It is worth to point out that we also implemented and tested a \textit{naive} hybrid A* algorithm \citep{Kurzer1057261}. This approach was chosen because it ensures that the robot reaches its final position with a specific orientation, which is important given that the limited work zone of the robotic arm. In our implementation, the search was performed on the entire map (modeled as an occupancy grid), where the crop rows were treated as occupied cells. However, this approach is considered \textit{naive} because the search is conducted across the entire grid, making it computationally expensive, especially for large, high-resolution maps. For example, the map used in our experiments covered 2.67 hectares and, at a resolution of 5 cm/pixel, contained 13,220,388 cells, causing the search algorithm to take an impractically long time to find a solution.

\subsection{Experiments Setup}

We evaluated these three path planning algorithms using a consistent set of 10000 randomly generated initial robot positions, orientations, and goal locations within a field comprising 65 crop rows and 207\,m corridor length, reflecting real-world conditions in Iowa State University Agronomy farm. Performance was measured across three key metrics: planning time, success rate, and average path length, as presented in Table \ref{tab:tab1}.
\subsection{Simulation Deployment}
To deploy the proposed path planning algorithms on a robot in real agricultural fields, we developed a pipeline that integrates the path planning output with the robot controller. First, we performed experiments in the simulation environment, constructed using the Gazebo simulator and the Cropcraft tool from \cite{cropcraft}. To ensure compatibility across all path planning algorithms, we standardized their output as a sequence of actions in the format $[Orientation, Move\_action]$. Fig. \ref{fig:sim} shows an example deployment of the pipeline in simulation with input action sequence as $[[0,0], [1,7], [1,0]]$. 

The waypoints for each corridor are generated and aligned with the centerline between crop rows based on the crop field dimensions and inter-row spacing. Since the simulated crop rows are planted in parallel, mirroring the structure of real agricultural fields, this ensures accurate waypoint placement. As mentioned in Section \ref{sec:RL}, movement action indicates three maneuvers: 0: Forward, 1:Backward, and \(\geq 2\): Switching rows. With the orientation values, navigation consists of up to three phases: exiting the current corridor, switching to the target corridor, and entering the target corridor, while the goal may be reached without leaving the current corridor.

From the movement action output of the algorithms, the pipeline extracts the current corridor waypoints and adjusts it (flipping if necessary) depending on whether the movement action value indicates forward or backward movement (0: Forward; 1: Backward). The MPC controller is then used to track the waypoints. When the value of the movement action indicates a row-switching maneuver (\(\geq 2\)), the PID controller executes the maneuver using the method proposed by \citep{ruiji2025}, rotating the robot to the desired orientation based on the input actions. 

\section{Discussion} 
\begin{figure}[t]
    \centering
    \includegraphics[width =0.45\textwidth]{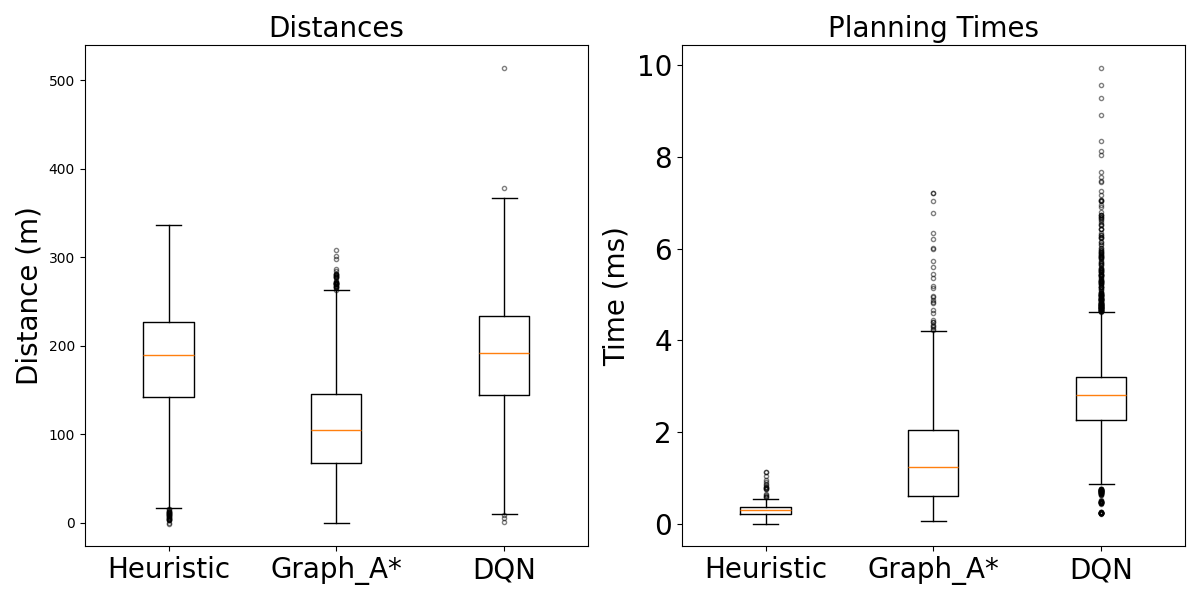}
    \caption{Boxplots comparing the performance of three path-planning algorithms (Heuristic, Graph A*, and DQN) in terms of (a) distance traveled and (b) planning time. }
    \label{fig:box}
\end{figure}
Table \ref{tab:tab1} shows that the Heuristic search method achieves the shortest planning time (0.28\,ms) and highest success rate (100\%), making it the most efficient and reliable path planning approach in this study. The Graph search method, though slightly slower (1.40\,ms), has only a minor failure rate (99.13\%). The DQN model requires a longer planning time (2.78\,ms) and has the lowest success rate (96.33\%). In addition, in the box plot shown in Fig. \ref{fig:box}, all three methods show distinct median values with symmetrically distributed whiskers, indicating data without significant skewness. Notably, both A* and DQN display several outliers extending above their upper whiskers, indicating occasional spikes in computation time.  

Overall, these results highlight the trade-offs among planning approaches. The Heuristic and Graph search methods consistently generate the paths with high reliability, while the DQN model, despite its learning capability, suffers from longer planning times and suboptimal success rate. This suggests that while learning-based methods offer adaptability, they require further refinement to improve efficiency and reliability, particularly in structured environments like crop fields.

In addition, although the trained DQN model with a large field can be applied directly to the smaller field, it fails and requires significantly longer time training on larger fields with more crop rows. The DQN model also requires consistent corridor length (10 units) for training, causing additional procedures for subsequent deployment. In contrast, both the Heuristic and Graph search methods can be applied to any field dimension with predefined initial robot positions, orientations, and goal positions. While the planning time for the Graph search method gradually increases with the field size, the planning time for the Heuristic method remains constant.
\section{Conclusion}
In this paper, we introduce an Unmanned Aerial System (UAS) survey pipeline for generating orthomosaic maps of the target crop field work zone and evaluate three path-planning strategies for autonomous nitrate sampling in large agricultural fields with structured crop rows: Heuristic search, Graph search (A*), and a DQN-based reinforcement learning approach. The Heuristic method outperformed the others, achieving the fastest planning time and perfect success rate. The Graph search method demonstrated near-optimal reliability and efficiency. In contrast, the DQN model struggled to balance adaptability with computational efficiency.

These findings highlight the effectiveness of rule-based methods (Heuristic and Graph search (A*)) in geometrically constrained agricultural environments, where regularity supports deterministic planning. Although learning-based approaches show promise for dynamic scenarios, their current limitations in reliability and scalability require further refinement. Future work could focus on automating waypoint extraction, developing hybrid path planning strategies, or refining learning-based approaches for dynamic scenarios, further advancing autonomous navigation in precision agriculture.

\begin{ack}
We thank Iowa State University for providing field research opportunities at the Curtiss farm and the Agronomy farm. This work was supported by: NSF/USDA-NIFA AIIRA AI Research Institute 2021-67021-35329..
\end{ack}

\bibliography{ifacconf}             
                                                   







\end{document}